\pdfoutput=1
\documentclass[11pt,a4paper]{article}
\usepackage[hyperref]{acl2019}
\usepackage{times}
\usepackage{latexsym}

\usepackage{url}

\aclfinalcopy 

\title{A computational linguistic study of personal recovery in bipolar disorder}

\author{Glorianna Jagfeld \\
  Spectrum Centre for Mental Health Research \\
  Lancaster University\\
  United Kingdom \\
  \texttt{g.jagfeld@lancaster.ac.uk}}

\date{}

\begin{document}
\maketitle

\begin{abstract}
    Mental health research can benefit increasingly fruitfully from computational linguistics methods, given the abundant availability of language data in the internet and advances of computational tools.
    This interdisciplinary 
    project will collect and analyse social media data of individuals diagnosed with bipolar disorder with regard to their recovery experiences.
    Personal recovery - living a satisfying and contributing life along symptoms of severe mental health issues - so far has only been investigated qualitatively with structured interviews and quantitatively with standardised questionnaires with mainly English-speaking participants in Western countries.
    Complementary to this evidence, computational linguistic methods allow us to analyse
    first-person accounts shared online in large quantities, representing unstructured settings and a more heterogeneous, multilingual population, 
    to draw a more complete picture of the aspects and mechanisms of personal recovery in bipolar disorder.
\end{abstract}

\section{Introduction and background}
Recent years have witnessed increased performance in many computational linguistics tasks such as syntactic and semantic parsing~\cite{NLPfromScratch_Collobert_11,CoNLLST_Zeman_18}, emotion classification~\cite{MultilingualEmotionClassification_Becker_17}, and sentiment analysis~\cite{SoASentimentSoADatasets_Barnes_17,BilingualSentimentEmbeddings_Barnes_18,DomainAdaptationSentiment_Barnes_18}, especially concerning the applicability of such tools to 
noisy online data.
Moreover, the field has made substantial progress in developing multilingual models and extending semantic annotation resources to languages beyond English~\cite{MultiWordNet_Pianta_02,MultilingualFrameNet_Boas_09,USAS_Multilingual_Piao_16,DutchLIWC_Boot_17}.

Concurrently, it has been argued for mental health research that it would constitute a \lq valuable critical step\rq~\cite{RecoverySystematicReview_Stuart_17} to analyse first-hand accounts by individuals with lived experience of severe mental health issues in blog posts, tweets, and discussion forums.
Several severe mental health difficulties, e.g., bipolar disorder~(BD) and schizophrenia are considered as chronic and clinical recovery, defined as being relapse and symptom free 
for a sustained period of time~\cite{RecoveryOlenzapineBD_Chengappa_05}, is considered difficult to achieve~\cite{BD_Forster_14,NICEGuideline_14,Schizophrenia_NIMH_16}.
Moreover, clinically recovered individuals often do not regain full social and educational/vocational functioning~\cite{OutcomeAffectivePsychosis_Strakowski_98,FirstEpisodeManiaStudy_Tohen_03}.
Therefore, research originating from initiatives by people with lived experience of mental health issues has been advocating emphasis on the individual's goals in recovery ~\cite{RecoveryRehabilitation_Deegan_88,RecoveryMentalIllness_Anthony_93}.
This movement gave rise to the concept of personal recovery~\cite{PsychologicalRecovery_Andresen_11,EvidenceBasedPsychiatryChange_Os_19}, loosely defined as a \lq way of living a satisfying, hopeful, and contributing life even with limitations caused by illness\rq~\cite{RecoveryMentalIllness_Anthony_93}.
The aspects of personal recovery have been conceptualised in various ways~\cite{ExploringRecovery_Young_99,RecoveryBipolarIQualitative_Mansell_10,PsychologicalApproachesRecoveryBD_Morrison_16}.
According to the frequently used CHIME model~\cite{ConceptualFrameworkRecoverySystematicReview_Leamy_11}, its main components are Connectedness, Hope and optimism, Identity, Meaning and purpose, and Empowerment. 

Here, we focus on BD, which is characterised by recurring episodes of depressed and elated (hypomanic or manic) mood~\cite{UnderstandingBDReport_Jones_10,BD_Forster_14}.
Bipolar spectrum disorders were estimated to affect approximately 2\% of the UK population~\cite{NICEGuideline_14} with rates ranging from 0.1\%-4.4\% across 11 other European, American and Asian countries~\cite{PrevalenceBP_Merikangas_11}.
Moreover, BD is associated with a high risk of suicide~\cite{SuicideAttemptsBP_Novick_10}, making its prevention and treatment important tasks for society.
BD-specific personal recovery research is motivated by mainly two facts:
First, the pole of positive/elevated mood and ongoing mood instability constitute core features of BD and pose special challenges compared to other mental health issues, such as unipolar depression~\cite{UnderstandingBDReport_Jones_10}.
Second, unlike for some other severe mental health difficulties, return to normal functioning is achievable 
given appropriate treatment~\cite{BipolarOutcome_Coryell_98,FirstEpisodeManiaStudy_Tohen_03,OutcomeBipolar_Goldberg_04}.

A substantial body of qualitative and quantitative research has shown the importance of personal recovery for individuals diagnosed with BD~\cite{RecoveryBipolarIQualitative_Mansell_10,UnderstandingBDReport_Jones_10,BipolarRecoveryQuestionnaire_Jones_13,RecoveryCBTPilotTrial_Jones_15,PsychologicalApproachesRecoveryBD_Morrison_16}.
Qualitative evidence mainly comes from \mbox{(semi-)}structured interviews and focus groups and 
has been criticised 
for small numbers of participants~\cite{RecoverySystematicReview_Stuart_17}
, lacking complementary quantitative evidence from larger samples~\cite{InternationalDifferencesRecoverySystematicReview_Slade_12}.
Some quantitative evidence stems from the standardised bipolar recovery questionnaire~\cite{BipolarRecoveryQuestionnaire_Jones_13} and a randomised control trial for recovery-focused cognitive-behavioural therapy~\cite{RecoveryCBTPilotTrial_Jones_15}.
Critically, previous research has taken place only 
in structured settings. 

What is more, the recovery concept emerged from research primarily conducted in English-speaking countries, 
mainly involving researchers and participants of Western ethnicity.
This might have led to a lack of non-Western notions of wellbeing in the concept, such as those found in indigenous 
peoples~\cite{InternationalDifferencesRecoverySystematicReview_Slade_12}, limiting its the applicability to a general population.
Indeed, the variation in BD prevalence rates from 0.1\% in India to 4.4\% in the US is striking. 
It has been shown that culture is an important factor in the diagnosis of BD~\cite{CultureManicSymptoms_Makin_06}, as well as on the causes attributed to mental health difficulties in general and treatments considered appropriate~\cite{TransculturalBD_Sanches_04,DepressionAcrossCultures_Chentsova-Dutton_15}.
While approaches to mental health classification from texts have long ignored the cultural dimension~\cite{CrossCulturalDifferencesDepressionLanguage_Loveys_18}, first studies show that online language of individuals affected by depression or related mental health difficulties differs significantly across cultures ~\cite{GenderCultureMentalHealthOnline_DeChoudhury_17,CrossCulturalDifferencesDepressionLanguage_Loveys_18}.

Hence, it seems timely to take into account the wealth of accounts of mental health difficulties and recovery stories from individuals of diverse ethnic and cultural backgrounds that are available in a multitude of languages on the internet. 
Corpus and computational linguistic methods are explicitly designed for processing large amounts of linguistic data~\cite{SpeechAndLanguageProcessing_Jurafsky_09,RoutledgeHandbookCorpusLinguistics_OKeeffe_10,CorpusLinguistics_McEnery_11,ComputationalToolsMethodsCorpus_Rayson_15}, and as discussed above, recent advances have made it feasible to apply them to noisy user-generated texts from diverse domains, including mental health~\cite{CLPsych1_Resnik_14,MultitaskMentalHealthSocialMedia_Benton_17}.
Computer-aided analysis of public social media data enables us to address several shortcomings in the 
scientific underpinning of personal recovery in BD by overcoming the small sample sizes of lab-collected data and including accounts from a more heterogeneous population.

In sum, our research questions are as follows:
(1) How is personal recovery discussed online by individuals meeting criteria for BD?
(2) What new insights do we get about personal recovery and factors that facilitate or hinder it?
We will investigate these questions in two parts, looking at English-language data by westerners and at multilingual data by individuals of diverse ethnicities.

\section{Data}
Previous work in computational linguistics and clinical psychology has tended to focus on the detection of mental health issues as classification tasks~\cite{DetectingMentalHealthFromLanguage_Arseniev-Koehler_18}.
Datasets have been collected for various conditions including BD using publicly available social-media data from Twitter~\cite{ADHDToSAD_Coppersmith_15} and 
Reddit~\citep{BDPrediction_Sekulic_18,SMHD_Cohan_18}
.
Unfortunately, the Twitter dataset is unavailable for further research.\footnote{Email communication with the first author of ~\citet{ADHDToSAD_Coppersmith_15}.}
In both Reddit datasets, mental health-related content was deliberately removed.
This allows the training of classifiers that try to predict the mental health of authors from excerpts that do not explicitly address mental health, yet it renders the data useless for analyses on how mental health is talked about online.
Due to this lack of appropriate existing publicly accessible datasets, we will create such resources and make them available to subsequent researchers.

We plan to collect data relevant for BD in general as well as for personal recovery in BD from three sources varying in their available amount versus depth of the accounts we expect to find: 1)~Twitter, 2)~Reddit (focusing on mental health-related content unlike previous work), 3)~
blogs authored by affected individuals.
Twitter and Reddit users with a BD diagnosis will be identified automatically via self-reported diagnosis statements, such
as \lq I was diagnosed with BD-I last week\rq.
To do so, we will extend on the diagnosis patterns and terms for BD  provided by~\citet{SMHD_Cohan_18}\footnote{\url{http://ir.cs.georgetown.edu/data/smhd/}}.
Implicit consent is assumed from users on these platforms to use their public tweets and posts.\textsuperscript{\ref{footnote:ethics}} 
Relevant blogs will be manually identified, and their authors will be contacted to obtain informed consent for using their texts.

Since language and culture are important factors in our research questions, we need information on the language of the texts and the country of residence of their authors\footnote{See Section~\ref{sec:ethics} for ethical considerations on this.\label{footnote:ethics}}, 
which is not provided in a structured format in the three data sources.
For language identification, Twitter employs an automatic tool~\cite{TwitterLanguageIdentification_15}, which can be used to filter tweets according to 60 language codes, and there are free, fairly accurate tools such as the Google Compact Language Detector\footnote{\url{https://github.com/CLD2Owners/cld2}}, which can be applied to Reddit and blog posts.
The location of Twitter users can be automatically inferred from their tweets~\cite{LocatingTwitterUsers_Cheng_10} or the (albeit noisy) location field in their user profiles~\cite{TweetLocation_Hecht_11}.
Only one attempt to classify the location of Reddit users has been published so far~\cite{GeocodingReddit_Harrigian_18} showing meagre results, indicating that the development of robust location classification approaches on this platform would constitute a valuable contribution. 

Some companies collect mental health-related online data and make them available to researchers subject to approval of their internal review boards, e.g., OurDataHelps\footnote{\url{https://ourdatahelps.org/}} by Qntfy or the peer-support forum provider 7~Cups\footnote{\url{https://7cups.com/}}.
Unlike `raw' social media data, these datasets have richer user-provided metadata and explicit consent for research usage.
On the other hand, less data is available
, the process to obtain access might be tedious within the short timeline of a PhD project and it might be impossible to share the used portions of the data with other researchers. 
Therefore, we will follow up the possibilities of obtaining access to these datasets, but in parallel also collect our own datasets to avoid dependence on external data providers.

\section{Methodology and Resources}
As explained in the introduction, the overarching aim of this project is to investigate in how far information conveyed in social media posts 
can complement more traditional research methods in clinical psychology to get insights into the recovery experience of individuals with a BD diagnosis.
Therefore, we will first conduct a systematic literature review of qualitative evidence 
to establish a solid base of what is already known about personal recovery experiences in BD for the subsequent social media studies.

Our research questions, which regard the experiences of different populations, lend themselves to 
several subprojects.
First, we will collect and analyse English-language data from westerners. 
Then, we will 
address ethnically diverse English-speaking populations and finally multilingual accounts.
This 
has the advantage that we can build data processing and methodological workflows along an increase in complexity of the data collection and analysis throughout the project.

In each project phase, we will employ a mixed-methods approach to combine the advantages of quantitative and qualitative methods~\cite{MixedMethods_Tashakkori_98,MixedMethods_Creswell_11}, 
which is established in mental health research~\cite{IntegratingQualiQuanti_Steckler_92,QualiQuanti_Baum_95,MixedMethodsHealthCare_Sale_02,CombiningQualiQuanti_Lund_12} and specifically recommended to investigate personal recovery~\cite{RecoveryResearchParadigms_Leonhardt_17}.
Quantitative methods are suitable to study observable behaviour such as language and yield more generalisable results by taking into account large samples.
However, they fall short of capturing the subjective, idiosyncratic meaning of socially constructed reality, which is important when studying individuals' recovery experience~\cite{StayingWellWithBD_Russell_05,RecoveryBipolarIQualitative_Mansell_10,PsychologicalApproachesRecoveryBD_Morrison_16,StayingWellBD_Crowe_18}.
Therefore, we will apply an explanatory sequential research design~\cite{MixedMethods_Creswell_11}
, starting with statistical analysis of the full dataset followed by a 
manual investigation of fewer examples, similar to \lq distant reading\rq~\cite{DistantReading_Moretti_13} in digital humanities.

Since previous research mainly employed (semi-)structured interviews and we do not expect to necessarily find the same aspects emphasised in unstructured settings, even less so when looking at a more diverse and non-English speaking population, we will not derive hypotheses from existing recovery models for testing on the online data.
Instead, we will start off with exploratory quantitative research using comparative analysis tools such as Wmatrix~\cite{KeyWordsKeySemanticDomainsWMatrix_Rayson_08} to uncover important linguistic features, e.g., on keywords and key concepts that occur with unexpected frequency in our collected datasets relative to reference corpora.
The underlying assumption is that keywords and key concepts are indicative of certain aspects of personal recovery, such as those specified in the CHIME model~\cite{ConceptualFrameworkRecoverySystematicReview_Leamy_11}, other previous 
research~\cite{RecoveryBipolarIQualitative_Mansell_10,PsychologicalApproachesRecoveryBD_Morrison_16,StayingWellBD_Crowe_18}, or novel ones.
Comparing online sources with transcripts of structured interviews or subcorpora originating from different cultural backgrounds might uncover aspects that were not prominently represented in the accounts studied in prior research.

A specific challenge will be to narrow down the data to parts relevant for personal recovery, since there is no control over the discussed topics compared to structured interviews. 
To investigate how individuals discuss personal recovery online and what (potentially unrecorded) aspects they associate with it, without a priori narrowing down the search-space to specific known keywords seems like a chicken-and-egg problem.
We propose to address this challenge by an iterative approach similar to the one taken in a corpus linguistic study of cancer metaphors~\cite{MetaphorCancer_Semino_17}.
Drawing on results from previous qualitative research~\cite{ConceptualFrameworkRecoverySystematicReview_Leamy_11,PsychologicalApproachesRecoveryBD_Morrison_16}, we will compile an initial dictionary of recovery-related terms.
Next, we will examine a small portion of the dataset manually, which will be partly randomly sampled and partly selected to contain recovery-related terms.
Based on this, we will be able to expand the dictionary and additionally automatically annotate semantic concepts of the identified relevant text passages using a semantic tagging approach such as the UCREL Semantic Analysis System (USAS)~\cite{USAS_Rayson_04}. Crucially for the multilingual aspect of the project
, USAS can tag semantic categories in eight languages~\cite{USAS_Multilingual_Piao_16}.
Then, semantic tagging will be applied to the full corpus to retrieve all text passages mentioning relevant concepts.
Furthermore, distributional semantics methods~\cite{DistributionalSemantics_Lenci_08,FrequencyToMeaning_Turney_10} can be used to find terms that frequently co-occur with words from our keyword dictionary. 
Occurrences of the identified keywords or concepts can be quantified in the full corpus to identify the importance 
of the related personal recovery aspects.

Linguistic Inquiry and Word Count~(LIWC)~\cite{LIWC_Pennebaker_15} is a frequently used tool in social-science text analysis to analyse emotional and cognitive 
components of texts and derive features for classification models~\cite{SMHD_Cohan_18,BDPrediction_Sekulic_18,DepressionMultiLab_Tackman_18, MeasuringSupportOnlineCommunities_Wang_18}.
LIWC counts target words organised in a manually constructed hierarchical dictionary without contextual disambiguation in the texts under analysis and has been psychometrically validated and developed 
for English exclusively.
While translations for several languages exist, e.g., Dutch~\cite{DutchLIWC_Boot_17},
and it is questionable to what extent LIWC concepts can be transferred to other languages and cultures by mere translation. We therefore aim to apply and develop methods that require less 
manual labour and are applicable to many 
languages and cultures.
One option constitute unsupervised methods, such as topic modelling, which has been applied to explore cultural differences in mental-health related online data already~\cite{GenderCultureMentalHealthOnline_DeChoudhury_17,CrossCulturalDifferencesDepressionLanguage_Loveys_18}.
The Differential Language Analysis ToolKit (DLATK)~\cite{DLATK_Schwartz_17} facilitates social-scientific language analyses, including tools for preprocessing, such as emoticon-aware tokenisers, filtering according to meta data, and analysis, e.g. via robust topic modelling methods.

Furthermore, emotion and sentiment analysis constitute useful tools to investigate the emotions involved in talking about recovery and identify factors that facilitate or hinder it.
There are many annotated datasets 
to train supervised classifiers~\cite{EmotionCorpora_Bostan_18,SoASentimentSoADatasets_Barnes_17} for these actively researched NLP tasks.
Machine learning methods were found to usually outperform rule-based approaches based on look-ups in dictionaries such as LIWC. Again, most annotated resources are English, but state of the art approaches
based on multilingual embeddings allow transferring models 
between languages~\cite{BilingualSentimentEmbeddings_Barnes_18}.

\section{Ethical considerations}
\label{sec:ethics}
Ethical considerations are established as essential part in planning mental health research and most research projects undergo approval by an ethics committee.
On the contrary, the computational linguistics community has started only recently to consider ethical questions~\cite{SocialImpactNLP_Hovy_16,EthicsNLPWorkshop_Hovy_17}.
Likely, this is because computational linguistics was traditionally concerned with publicly available, impersonal texts such as newspapers or texts published with some temporal distance
, which left a distance between the text and author.
Conversely, recent social media research often deals with highly personal information of living individuals, who can be directly affected by the outcomes~\cite{SocialImpactNLP_Hovy_16}.

\citet{SocialImpactNLP_Hovy_16} discuss issues that can arise when constructing datasets from social media and conducting analyses or developing predictive models based on these data, which we review here in relation to our project:
Demographic bias in sampling the data can lead to exclusion of minority groups, resulting in overgeneralisation of models based on these data.
As discussed in the introduction, personal recovery research suffers from 
a bias towards English-speaking Western individuals of white ethnicity.
By studying multilingual accounts of 
ethnically diverse populations we explicitly address the demographic bias of previous research.
Topic overexposure is tricky to address, where certain groups are perceived as abnormal when research repeatedly finds that their language is different or more difficult to process.
Unlike previous research~\cite{ADHDToSAD_Coppersmith_15,SMHD_Cohan_18,BDPrediction_Sekulic_18} our goal is not to reveal particularities in the language of individuals affected by mental health problems.
Instead, we will compare accounts of individuals with BD from different settings (structured interviews versus informal online discourse) and of different 
backgrounds.
While the latter 
bears the risk to overexpose certain minority groups, we will pay special attention to this in the dissemination of our results.

Lastly, most research, even when conducted with the best intentions, suffers from the dual-use problem~\cite{ImperativeResponsibility_Jonas_84}, in that it can be misused or have consequences that affect people's life negatively.
For this reason, we refrain from publishing mental health classification methods, which could be used, for example, by health insurance companies for the risk assessment of applicants based on their social media profiles.

If and how informed consent needs to be obtained for research on social media data is a debated issue~\cite{EthicalQualitativeInternetResearch_Eysenbach_01, ResearchSocialMediaUsersViews_Beninger_14, SocialMonitoringPublicHealth_Paul_17}, mainly because it is not straightforward to determine if posts are made in a public or private context.
From a legal point of view, the privacy policies of Twitter\footnote{\url{https://cdn.cms-twdigitalassets.com/content/dam/legal-twitter/site-assets/privacy-policy-new/Privacy-Policy-Terms-of-Service_EN.pdf}} and Reddit\footnote{\url{www.redditinc.com/policies/privacy-policy}}, explicitly allow analysis of the user contents by third party, but it is unclear to what extent users are aware of this when posting to these platforms~\cite{EthicsOnlineResearch_Ch4_TwitterDataSource_Ahmed_17}.
However, in practice it is often infeasible to seek retrospective consent from hundreds or thousands of social media users. According to current ethical guidelines for social media research~\cite{EthicalMentalHealthSocialMediaResearch_Benton_17, EthicalFrameworkPublishingTwitterData_Williams_17} and practice in comparable research projects~\cite{SuicidalityTwitter_ODea_15, EthicsOnlineResearch_Ch4_TwitterDataSource_Ahmed_17}, it is regarded as acceptable to waive explicit consent if the anonymity of the users is preserved.
Therefore, we will not ask the account holders of Twitter and Reddit posts included in our datasets for their consent.

\citet{EthicalMentalHealthSocialMediaResearch_Benton_17} formulate guidelines for ethical social media health research that pertain especially to data collection and sharing.
In line with these, we will only share anonymised and paraphrased excerpts from the texts, as it is often possible to recover a user name via a web search for the verbatim text of a post.
However, we will make the original texts available as datasets to subsequent research under a data usage agreement. 
Since the (automatic) annotation of demographic variables in parts of our dataset constitutes especially sensitive information on minority status in conjunction with mental health, we will only share these annotations with researchers that demonstrate a genuine need for them, i.e. to verify our results or to investigate certain research questions.

Another important question is in which situations of encountering content indicative of a risk of self-harm or harm to others it would be appropriate or even required by duty of care for the research team to pass on information to 
authorities.
Surprisingly, we could only find two mentions of this issue in social media research~\citep{SuicidalityTwitter_ODea_15,SuicdeOnlineStudy_Young_18}.
Acknowledging that suicidal ideation fluctuates~\citep{TrajectoriesSuicidalIdeation_Prinstein_08}
, we accord with the ethical review board's requirement in~\citet{SuicidalityTwitter_ODea_15} to only analyse content posted at least three months ago.
If the research team, which includes clinical psychologists, still perceives users at risk 
we will make use of the reporting facilities of Twitter and Reddit.

As a central component we consider the 
involvement of individuals with lived experience in 
our project, an aspect which is missing in the discussion of ethical social media health research so far.
The proposal has been presented to an advisory board of individuals with a BD diagnosis and was received positively.
The advisory board will be consulted at several stages of the project to inform the research design, analysis, and publication of results.
We believe that board members can help to address several of the raised ethical problems, e.g.,  
shaping the research questions to avoid feeding into existing biases or overexposing certain groups and highlighting potentially harmful interpretations and uses of our results. 

\section{Impact and conclusion}
The importance of the recovery concept in the design of mental health services has recently been prominently reinforced, suggesting ‘recovery-oriented social enterprises as key component of the integrated service’~\cite{EvidenceBasedPsychiatryChange_Os_19}. We think that a recovery approach as leading principle for national or global health service strategies, should be informed by voices of individuals as diverse as those it is supposed to serve. Therefore, we expect the proposed investigations of views on recovery by previously under-researched ethnic, language, and cultural groups to yield valuable insights on the appropriateness of the recovery approach for a wider population.
The datasets collected in this project can serve as useful resources for future research.
More generally, our social-media data-driven approach could be applied to investigate other 
areas of mental health if it proves successful in leading to relevant new insights.
 
Finally, this project is an interdisciplinary endeavour, combining clinical psychology, input from individuals with lived experience of BD, and computational linguistics. While this comes with the challenges of cross-disciplinary research, it has the potential to apply and develop state-of-the-art NLP methods in a way that is psychologically and ethically sound as well as informed and approved by affected people to increase our knowledge of severe mental illnesses such as BD.

\section*{Acknowledgments}
I would like to thank my supervisors Steven Jones, Fiona Lobban, and Paul Rayson for their guidance in this project.
My heartfelt thanks go also to Chris Lodge, service user researcher at the Spectrum Centre, and the members of the advisory panel he coordinates that offer feedback on this project based on their lived experience of BD.
Further, I would like to thank Masoud Rouhizadeh for his helpful comments during pre-submission mentoring and the anonymous reviewers.
This project is funded by the Faculty of Health and Medicine at Lancaster University as part of a doctoral scholarship.

\bibliography{library}

\begin{thebibliography}{84}
\expandafter\ifx\csname natexlab\endcsname\relax\def\natexlab#1{#1}\fi

\bibitem[{Ahmed et~al.(2017)Ahmed, Bath, and
  Demartini}]{EthicsOnlineResearch_Ch4_TwitterDataSource_Ahmed_17}
Wasim Ahmed, Peter~A. Bath, and Gianluca Demartini. 2017.
\newblock \href {https://doi.org/10.1016/j.jocn.2005.03.017} {{Using Twitter as
  a data source: an overview of ethical, legal and methodological challenges}}.
\newblock In Kandy Woodfield, editor, \emph{The Ethics of Online Research},
  pages 79--107. Emerald Books.

\bibitem[{Andresen et~al.(2011)Andresen, Caputi, and
  Oades}]{PsychologicalRecovery_Andresen_11}
Retta Andresen, Peter Caputi, and Lindsay~G Oades. 2011.
\newblock \href {https://doi.org/10.1002/9781119975182} {\emph{{Psychological
  Recovery: Beyond Mental Illness}}}.
\newblock John Wiley {\&} Sons, Ltd, Chichester, West Sussex.

\bibitem[{Anthony(1993)}]{RecoveryMentalIllness_Anthony_93}
William~A. Anthony. 1993.
\newblock {Recovery from mental illness: the guiding vision of the mental
  health system in the 1990s}.
\newblock \emph{Psychosocial Rehabilitation Journal}, 16(4):11--23.

\bibitem[{Arseniev-Koehler et~al.(2018)Arseniev-Koehler, Mozgai, and
  Scherer}]{DetectingMentalHealthFromLanguage_Arseniev-Koehler_18}
Alina Arseniev-Koehler, Sharon Mozgai, and Stefan Scherer. 2018.
\newblock \href {https://doi.org/10.18653/v1/W18-0601} {{What type of happiness
  are you looking for? - A closer look at detecting mental health from
  language}}.
\newblock In \emph{Proceedings of the Fifth Workshop on Computational
  Linguistics and Clinical Psychology: From Keyboard to Clinic}, pages 1--12.

\bibitem[{Barnes et~al.(2017)Barnes, Klinger, and Schulte~im
  Walde}]{SoASentimentSoADatasets_Barnes_17}
Jeremy Barnes, Roman Klinger, and Sabine Schulte~im Walde. 2017.
\newblock \href {https://doi.org/10.18653/V1/W17-5202} {{Assessing
  State-of-the-Art Sentiment Models on State-of-the-Art Sentiment Datasets}}.
\newblock In \emph{Proceedings of the 8th Workshop on Computational Approaches
  to Subjectivity, Sentiment and Social Media Analysis}, pages 2--12.

\bibitem[{Barnes et~al.(2018{\natexlab{a}})Barnes, Klinger, and Schulte~im
  Walde}]{BilingualSentimentEmbeddings_Barnes_18}
Jeremy Barnes, Roman Klinger, and Sabine Schulte~im Walde. 2018{\natexlab{a}}.
\newblock \href {http://arxiv.org/abs/1805.09016
  http://aclweb.org/anthology/C18-1070} {{Bilingual Sentiment Embeddings: Joint
  Projection of Sentiment Across Languages}}.
\newblock In \emph{Proceedings of the 56th Annual Meeting of the Association
  for Computational Linguistics (ACL)}, pages 2483--2493, Melbourne.

\bibitem[{Barnes et~al.(2018{\natexlab{b}})Barnes, Klinger, and
  Walde}]{DomainAdaptationSentiment_Barnes_18}
Jeremy Barnes, Roman Klinger, and Sabine Schulte~im Walde. 2018{\natexlab{b}}.
\newblock \href {http://arxiv.org/abs/1806.04381} {{Projecting Embeddings for
  Domain Adaptation: Joint Modeling of Sentiment Analysis in Diverse Domains}}.
\newblock In \emph{Proceedings of the 27th International Conference on
  Computational Linguistics}, pages 818--830.

\bibitem[{Baum(1995)}]{QualiQuanti_Baum_95}
Frances Baum. 1995.
\newblock \href {https://doi.org/10.1016/0277-9536(94)E0103-Y} {{Researching
  public health: Behind the qualitative-quantitative methodological debate}}.
\newblock \emph{Social Science and Medicine}, 40(4):459--468.

\bibitem[{Becker et~al.(2017)Becker, Moreira, and dos
  Santos}]{MultilingualEmotionClassification_Becker_17}
Karin Becker, Viviane~P. Moreira, and Aline G.~L. dos Santos. 2017.
\newblock \href {https://doi.org/10.1016/j.ipm.2016.12.008} {{Multilingual
  emotion classification using supervised learning: comparative experiments}}.
\newblock \emph{Information Processing and Management}, 53(3):684--704.

\bibitem[{Beninger et~al.(2014)Beninger, Fry, Jago, Lepps, Nass, and
  Silvester}]{ResearchSocialMediaUsersViews_Beninger_14}
Kelsey Beninger, Alexandra Fry, Natalie Jago, Hayley Lepps, Laura Nass, and
  Hannah Silvester. 2014.
\newblock \href
  {http://www.natcen.ac.uk/media/282288/p0639-research-using-social-media-report-final-190214.pdf}
  {{Research using Social Media; Users' Views}}.

\bibitem[{Benton et~al.(2017{\natexlab{a}})Benton, Coppersmith, and
  Dredze}]{EthicalMentalHealthSocialMediaResearch_Benton_17}
Adrian Benton, Glen Coppersmith, and Mark Dredze. 2017{\natexlab{a}}.
\newblock \href {https://doi.org/10.18653/v1/W17-1612} {{Ethical Research
  Protocols for Social Media Health Research}}.
\newblock \emph{Proceedings of the First Workshop on Ethics in Natural Language
  Processing}, page 94–102.

\bibitem[{Benton et~al.(2017{\natexlab{b}})Benton, Mitchell, and
  Hovy}]{MultitaskMentalHealthSocialMedia_Benton_17}
Adrian Benton, Margaret Mitchell, and Dirk Hovy. 2017{\natexlab{b}}.
\newblock \href {https://doi.org/10.1890/06-0645.1} {{Multi-Task Learning for
  Mental Health using Social Media Text}}.
\newblock In \emph{Proceedings of the 15th Conference of the European Chapter
  of the Association for Computational Linguistics (EACL)}, volume~1, pages
  152--162.

\bibitem[{Boas(2009)}]{MultilingualFrameNet_Boas_09}
Hans~C. Boas, editor. 2009.
\newblock \href {https://doi.org/10.1093/ijl/ecp034} {\emph{{Multilingual
  FrameNets in Computational Lexicography: Methods and Applications.}}}
\newblock Mouton de Gruyter, Berlin.

\bibitem[{Boot et~al.(2017)Boot, Zijlstra, and Geenen}]{DutchLIWC_Boot_17}
Peter Boot, Hanna Zijlstra, and Rinie Geenen. 2017.
\newblock \href {https://doi.org/10.1075/dujal.6.1.04boo} {{The Dutch
  translation of the Linguistic Inquiry and Word Count (LIWC) 2007
  dictionary}}.
\newblock \emph{Dutch Journal of Applied Linguistics}, 6(1):65 -- 76.

\bibitem[{Bostan and Klinger(2018)}]{EmotionCorpora_Bostan_18}
Laura-Ana-Maria Ana~Maria Bostan and Roman Klinger. 2018.
\newblock \href {https://doi.org/10.17226/11340} {{An Analysis of Annotated
  Corpora for Emotion Classification in Text}}.
\newblock In \emph{Proceedings of the 27th International Conference on
  Computational Linguistics}, pages 2104--2119. Association for Computational
  Linguistics.

\bibitem[{Cheng et~al.(2010)Cheng, Caverlee, and
  Lee}]{LocatingTwitterUsers_Cheng_10}
Zhiyuan Cheng, James Caverlee, and Kyumin Lee. 2010.
\newblock \href {https://doi.org/10.1145/1871437.1871535} {{You are where you
  tweet: A content-based approach to geo-locating Twitter users}}.
\newblock \emph{Proceedings of the 19th ACM International Conference on
  Information and Knowledge Management}, pages 759--768.

\bibitem[{Chengappa et~al.(2005)Chengappa, Hennen, Baldessarini, Kupfer,
  Yatham, Gershon, Baker, and Tohen}]{RecoveryOlenzapineBD_Chengappa_05}
K.~N.~Roy Chengappa, John Hennen, Ross~J. Baldessarini, David~J. Kupfer,
  Lakshmi~N. Yatham, Samuel Gershon, Robert~W. Baker, and Mauricio Tohen. 2005.
\newblock \href {https://doi.org/10.1111/j.1399-5618.2004.00171.x} {{Recovery
  and functional outcomes following olanzapine treatment for bipolar I mania}}.
\newblock \emph{Bipolar Disorders}, 7(1):68--76.

\bibitem[{Chentsova-Dutton et~al.(2014)Chentsova-Dutton, Ryder, and
  Tsai}]{DepressionAcrossCultures_Chentsova-Dutton_15}
Yulia~E. Chentsova-Dutton, Andrew~G. Ryder, and Jeanne Tsai. 2014.
\newblock \href {https://doi.org/10.1017/CBO9781107415324.004} {{Understanding
  depression across cultural contexts}}.
\newblock In Ian~H. Gotlib and Constance~L. Hammen, editors, \emph{Handbook of
  Depression}, pages 337--354. Guilford Press.

\bibitem[{Cohan et~al.(2018)Cohan, Desmet, Macavaney, Yates, Soldaini,
  Macavaney, and Goharian}]{SMHD_Cohan_18}
Arman Cohan, Bart Desmet, Sean Macavaney, Andrew Yates, Luca Soldaini, Sean
  Macavaney, and Nazli Goharian. 2018.
\newblock \href {https://www.aclweb.org/anthology/C18-1126} {{SMHD: A
  Large-Scale Resource for Exploring Online Language Usage for Multiple Mental
  Health Conditions}}.
\newblock In \emph{Proceedings of the 27th International Conference on
  Computational Linguistics (COLING)}, pages 1485–--1497, Santa Fe.
  Association for Computational Linguistics.

\bibitem[{Collobert et~al.(2011)Collobert, Weston, Bottou, Karlen, Kavukcuoglu,
  and Kuksa}]{NLPfromScratch_Collobert_11}
Ronan Collobert, Jason Weston, Léon Bottou, Michael Karlen, Koray Kavukcuoglu,
  and Pavel Kuksa. 2011.
\newblock \href {https://doi.org/10.1.1.231.4614} {{Natural language processing
  (almost) from scratch}}.
\newblock \emph{Journal of Machine Learning Research}, 12:2493--2537.

\bibitem[{Coppersmith et~al.(2015)Coppersmith, Dredze, Harman, and
  Hollingshead}]{ADHDToSAD_Coppersmith_15}
Glen Coppersmith, Mark Dredze, Craig Harman, and Kristy Hollingshead. 2015.
\newblock \href {https://doi.org/10.1890/04-0298} {{From ADHD to SAD: Analyzing
  the Language of Mental Health on Twitter through Self-Reported Diagnoses}}.
\newblock In \emph{Conference of the North American Chapter of the Association
  for Computational Linguistics – Human Language Technologies (NAACL)}, pages
  1--10.

\bibitem[{Coryell et~al.(1998)Coryell, Turvey, Endicott, Leon, Mueller,
  Solomon, and Keller}]{BipolarOutcome_Coryell_98}
William Coryell, Carolyn Turvey, Jean Endicott, Andrew~C. Leon, Timothy
  Mueller, David Solomon, and Martin Keller. 1998.
\newblock \href {https://doi.org/10.1016/S0165-0327(98)00043-3} {{Bipolar I
  affective disorder: Predictors of outcome after 15 years}}.
\newblock \emph{Journal of Affective Disorders}, 50(2-3):109--116.

\bibitem[{Creswell and Plano~Clark(2011)}]{MixedMethods_Creswell_11}
John~W. Creswell and Vicki~L. Plano~Clark. 2011.
\newblock \href {https://books.google.ch/books?id=YcdlPWPJRBcC}
  {\emph{{Designing and Conducting Mixed Methods Research}}}.
\newblock SAGE Publications.

\bibitem[{Crowe and Inder(2018)}]{StayingWellBD_Crowe_18}
Marie Crowe and Maree Inder. 2018.
\newblock \href {https://doi.org/10.1111/jpm.12455} {{Staying well with bipolar
  disorder: A qualitative analysis of five-year follow-up interviews with young
  people}}.
\newblock \emph{Journal of Psychiatric and Mental Health Nursing},
  25(4):236--244.

\bibitem[{De~Choudhury et~al.(2017)De~Choudhury, Logar, Sharma, Eekhout, and
  Nielsen}]{GenderCultureMentalHealthOnline_DeChoudhury_17}
Munmun De~Choudhury, Tomaz Logar, Sanket~S. Sharma, Wouter Eekhout, and
  René~Clausen Nielsen. 2017.
\newblock \href {https://doi.org/10.1145/2998181.2998220} {{Gender and
  Cross-Cultural Differences in Social Media Disclosures of Mental Illness}}.
\newblock In \emph{Proceedings of the 2017 ACM Conference on Computer Supported
  Cooperative Work and Social Computing}, pages 353--369.

\bibitem[{Deegan(1988)}]{RecoveryRehabilitation_Deegan_88}
Patricia~E. Deegan. 1988.
\newblock \href {https://doi.org/10.1037/h0099565} {{Recovery: The lived
  experience of rehabilitation.}}
\newblock \emph{Psychosocial Rehabilitation Journal}, 11(4):11--19.

\bibitem[{Eysenbach and
  Till(2001)}]{EthicalQualitativeInternetResearch_Eysenbach_01}
Gunther Eysenbach and James~E. Till. 2001.
\newblock \href {https://doi.org/10.1136/bmj.313.7055.438} {{Ethical issues in
  qualitative research on internet communities}}.
\newblock \emph{BMJ}, 323(7055):1103--1105.

\bibitem[{Forster(2014)}]{BD_Forster_14}
Peter Forster. 2014.
\newblock \href
  {https://doi.org/https://doi.org/10.1016/B978-0-12-385157-4.01077-0}
  {{Bipolar Disorder}}.
\newblock \emph{Encyclopedia of the Neurological Sciences}, pages 420--424.

\bibitem[{Goldberg and Harrow(2004)}]{OutcomeBipolar_Goldberg_04}
Joseph~F. Goldberg and Martin Harrow. 2004.
\newblock \href {https://doi.org/10.1016/S0165-0327(03)00161-7} {{Consistency
  of remission and outcome in bipolar and unipolar mood disorders: A 10-year
  prospective follow-up}}.
\newblock \emph{Journal of Affective Disorders}, 81(2):123--131.

\bibitem[{Harrigian(2018)}]{GeocodingReddit_Harrigian_18}
Keith Harrigian. 2018.
\newblock \href {https://praw.readthedocs.io/en/latest/} {{Geocoding without
  geotags: a text-based approach for reddit}}.
\newblock In \emph{Proceedings of the 2018 EMNLP Workshop W-NUT: The 4th
  Workshop on Noisy User-generated Text}, pages 17--27.

\bibitem[{Hecht et~al.(2011)Hecht, Hong, Suh, and Chi}]{TweetLocation_Hecht_11}
Brent Hecht, Lichan Hong, Bongwon Suh, and Ed~H. Chi. 2011.
\newblock \href {https://doi.org/10.1145/1978942.1978976} {{Tweets from Justin
  Bieber's heart: the dynamics of the location field in user profiles}}.
\newblock In \emph{Proceedings of the SIGCHI Conference on Human Factors in
  Computing Systems}, pages 237--246.

\bibitem[{Heylighen et~al.(2014)Heylighen, Neuckermans, Akazawa-Ogawa,
  Shichiri, Nishio, Yoshida, Niki, Hagihara, Dempsey, Gooding, Jones, Akers,
  Eaton, Tyler, Gatherer, Brabban, Long, Lobban, Jones, Apazoglou, K{\"{u}}ng,
  Cordera, Aubry, Dayer, Vuilleumier, Piguet, {Russell S.J.}, Jones, Cooke,
  Cooke, Falk, Marshal, Jones, Smith, Mulligan, Lobban, Law, Dunn, Welford,
  Kelly, Mulligan, Morrison, Tyler, Lobban, Sutton, Depp, Johnson, Laidlaw,
  Jones, Murray, Leitan, Thomas, Michalak, Johnson, Jones, Perich, Berk, Berk,
  Monk, Flaherty, Frank, Hoskinson, Kupfer, Spillane, Matvienko-Sikar, Larkin,
  Corcoran, and Arensman}]{NICEGuideline_14}
Ann Heylighen, Herman Neuckermans, Yoko Akazawa-Ogawa, Mototada Shichiri, Keiko
  Nishio, Yasukazu Yoshida, Etsuo Niki, Yoshihisa Hagihara, R.~C. Dempsey,
  P.~A. Gooding, Steven~Huntley Jones, Nadia Akers, Jayne Eaton, Elizabeth
  Tyler, Amanda Gatherer, Alison Brabban, Rita~Marie Long, Anne~Fiona Lobban,
  Raya~A. Jones, Kallia Apazoglou, Anne-Lise K{\"{u}}ng, Paolo Cordera,
  Jean-Michel Aubry, Alexandre Dayer, Patrik Vuilleumier, Camille Piguet,
  {Russell S.J.}, Prof~Steven Jones, Anne Cooke, Anne Cooke, Karin Falk,
  I.~Marshal, Steven~Huntley Jones, Gina Smith, Lee~D Mulligan, Fiona Lobban,
  Heather Law, Graham Dunn, Mary Welford, James Kelly, John Mulligan, Anthony~P
  Morrison, Elizabeth Tyler, Anne~Fiona Lobban, Chris Sutton, Colin Depp,
  Sheri~L Johnson, Ken Laidlaw, Steven~Huntley Jones, Greg Murray, Nuwan~D
  Leitan, Neil Thomas, Erin~E Michalak, Sheri~L Johnson, Steven~Huntley Jones,
  Tania Perich, Lesley Berk, Michael Berk, Timothy~H. Monk, Joseph~F. Flaherty,
  Ellen Frank, Kathleen Hoskinson, David~J. Kupfer, Ailbhe Spillane, Karen
  Matvienko-Sikar, Celine Larkin, Paul Corcoran, and Ella Arensman. 2014.
\newblock \href {https://doi.org/10.1016/j.cpr.2017.01.002} {\emph{{Bipolar
  disorder: assessment and management}}}, volume~7.
\newblock National Institute for Health and Care Excellence.

\bibitem[{Hovy et~al.(2017)Hovy, Spruit, Mitchell, Bender, Strube, and
  Wallach}]{EthicsNLPWorkshop_Hovy_17}
Dirk Hovy, Shannon Spruit, Margaret Mitchell, Emily~M Bender, Michael Strube,
  and Hanna Wallach. 2017.
\newblock \href {http://aclweb.org/anthology/W17-1600} {\emph{{Proceedings of
  the First ACL Workshop on Ethics in Natural Language Processing}}}.
\newblock Association for Computational Linguistics.

\bibitem[{Hovy and Spruit(2016)}]{SocialImpactNLP_Hovy_16}
Dirk Hovy and Shannon~L. Spruit. 2016.
\newblock \href {https://doi.org/10.18653/v1/P16-2096o} {{The Social Impact of
  Natural Language Processing}}.
\newblock In \emph{Proceedings of the 54th Annual Meeting of the Association
  for Computational Linguistics (ACL)}, pages 591--598.

\bibitem[{Jonas(1984)}]{ImperativeResponsibility_Jonas_84}
Hans Jonas. 1984.
\newblock \emph{{The Imperative of Responsibility: Foundations of an Ethics for
  the Technological Age}}.
\newblock University of Chicago Press, Chicago.

\bibitem[{Jones et~al.(2010)Jones, Lobban, and
  Cook}]{UnderstandingBDReport_Jones_10}
Steven Jones, Fiona Lobban, and Anne Cook. 2010.
\newblock \href {https://www1.bps.org.uk/system/files/Public files/cat-653.pdf}
  {\emph{{Understanding Bipolar Disorder - Why some people experience extreme
  mood states and what can help}}}.
\newblock British Psychological Society.

\bibitem[{Jones et~al.(2012)Jones, Mulligan, Higginson, Dunn, and
  Morrison}]{BipolarRecoveryQuestionnaire_Jones_13}
Steven Jones, Lee~D. Mulligan, Sally Higginson, Graham Dunn, and Anthony~P
  Morrison. 2012.
\newblock \href {https://doi.org/10.1016/j.jad.2012.10.003} {{The bipolar
  recovery questionnaire: psychometric properties of a quantitative measure of
  recovery experiences in bipolar disorder}}.
\newblock \emph{Journal of Affective Disorders}, 147(1-3):34--43.

\bibitem[{Jones et~al.(2015)Jones, Smith, Mulligan, Lobban, Law, Dunn, Welford,
  Kelly, Mulligan, and Morrison}]{RecoveryCBTPilotTrial_Jones_15}
Steven~H. Jones, Gina Smith, Lee~D. Mulligan, Fiona Lobban, Heather Law, Graham
  Dunn, Mary Welford, James Kelly, John Mulligan, and Anthony~P. Morrison.
  2015.
\newblock \href {https://doi.org/10.1192/bjp.bp.113.141259} {{Recovery-focused
  cognitive-behavioural therapy for recent-onset bipolar disorder: randomised
  controlled pilot trial}}.
\newblock \emph{British Journal of Psychiatry}, 206(1):58--66.

\bibitem[{Jurafsky and Martin(2009)}]{SpeechAndLanguageProcessing_Jurafsky_09}
Daniel Jurafsky and James~H. Martin. 2009.
\newblock \emph{{Speech and Language Processing (2nd Edition)}}.
\newblock Prentice-Hall, Inc., Upper Saddle River, USA.

\bibitem[{Leamy et~al.(2011)Leamy, Bird, Le~Boutillier, Williams, and
  Slade}]{ConceptualFrameworkRecoverySystematicReview_Leamy_11}
Mary Leamy, Victoria Bird, Clair Le~Boutillier, Julie Williams, and Mike Slade.
  2011.
\newblock \href {https://doi.org/10.1192/bjp.bp.110.083733} {{Conceptual
  framework for personal recovery in mental health: Systematic review and
  narrative synthesis}}.
\newblock \emph{British Journal of Psychiatry}, 199(6):445--452.

\bibitem[{Lenci(2008)}]{DistributionalSemantics_Lenci_08}
Alessandro Lenci. 2008.
\newblock \href {http://linguistica.sns.it/RdL/20.1/ALenci.pdf}
  {{Distributional semantics in linguistic and cognitive research}}.
\newblock \emph{Italian Journal of Linguistics}, 20(1):1--31.

\bibitem[{Leonhardt et~al.(2017)Leonhardt, Huling, Hamm, Roe, Hasson-Ohayon,
  McLeod, and Lysaker}]{RecoveryResearchParadigms_Leonhardt_17}
Bethany~L. Leonhardt, Kelsey Huling, Jay~A. Hamm, David Roe, Ilanit
  Hasson-Ohayon, Hamish~J. McLeod, and Paul~H. Lysaker. 2017.
\newblock \href {https://doi.org/10.1080/14737175.2017.1378099} {{Recovery and
  serious mental illness: a review of current clinical and research paradigms
  and future directions}}.
\newblock \emph{Expert Review of Neurotherapeutics}, 17(11):1117--1130.

\bibitem[{Loveys et~al.(2018)Loveys, Torrez, Fine, Moriarty, and
  Coppersmith}]{CrossCulturalDifferencesDepressionLanguage_Loveys_18}
Kate Loveys, Jonathan Torrez, Alex Fine, Glen Moriarty, and Glen Coppersmith.
  2018.
\newblock \href {https://doi.org/10.1002/jcu.10150} {{Cross-cultural
  differences in language markers of depression online}}.
\newblock In \emph{Proceedings of the Fifth Workshop on Computational
  Linguistics and Clinical Psychology: From Keyboard to Clinic}, pages 78--87.

\bibitem[{Lund(2012)}]{CombiningQualiQuanti_Lund_12}
Thorleif Lund. 2012.
\newblock \href {https://doi.org/10.1080/00313831.2011.568674} {{Combining
  Qualitative and Quantitative Approaches: Some Arguments for Mixed Methods
  Research}}.
\newblock \emph{Scandinavian Journal of Educational Research}, 56(2):155--165.

\bibitem[{Mackin et~al.(2006)Mackin, Targum, Kalali, Rom, and
  Young}]{CultureManicSymptoms_Makin_06}
Paul Mackin, Steven~D. Targum, Amir Kalali, Dror Rom, and Allan~H. Young. 2006.
\newblock \href {https://doi.org/10.1192/bjp.bp.105.013920} {{Culture and
  assessment of manic symptoms}}.
\newblock \emph{British Journal of Psychiatry}, 189(04):379--380.

\bibitem[{Mansell et~al.(2010)Mansell, Powell, Pedley, Thomas, and
  Jones}]{RecoveryBipolarIQualitative_Mansell_10}
Warren Mansell, Seth Powell, Rebecca Pedley, Nia Thomas, and Sarah~Amelia
  Jones. 2010.
\newblock \href {https://doi.org/10.1348/014466509X451447} {{The process of
  recovery from bipolar I disorder: A qualitative analysis of personal accounts
  in relation to an integrative cognitive model}}.
\newblock \emph{British Journal of Clinical Psychology}, 49(2):193--215.

\bibitem[{McEnery and Hardie(2011)}]{CorpusLinguistics_McEnery_11}
Tony McEnery and Andrew Hardie. 2011.
\newblock \href {https://books.google.co.uk/books?id=3j3Wn_ZT1qwC}
  {\emph{{Corpus Linguistics: Method, Theory and Practice}}}.
\newblock Cambridge Textbooks in Linguistics. Cambridge University Press.

\bibitem[{Merikangas et~al.(2011)Merikangas, Jin, He, Kessler, Lee, Sampson,
  Viana, Andrade, Hu, Karam, Ladea, Mora, Browne, Ono, Posada-Villa, Sagar, and
  Zarkov}]{PrevalenceBP_Merikangas_11}
Kathleen~R. Merikangas, Robert Jin, Jian-ping He, Ronald~C. Kessler, Sing Lee,
  Nancy~A. Sampson, Maria~Carmen Viana, Laura~Helena Andrade, Chiyi Hu, Elie~G.
  Karam, Maria Ladea, Maria Elena~Medina Mora, Mark~Oakley Browne, Yutaka Ono,
  Jose Posada-Villa, Rajesh Sagar, and Zahari Zarkov. 2011.
\newblock \href {https://doi.org/10.1001/archgenpsychiatry.2011.12}
  {{Prevalence and correlates of bipolar spectrum disorder in the world mental
  health survey initiative}}.
\newblock \emph{Archives of general psychiatry}, 68(3):241--251.

\bibitem[{Moretti(2013)}]{DistantReading_Moretti_13}
Franco Moretti. 2013.
\newblock \emph{{Distant reading}}.
\newblock Verso, London.

\bibitem[{Morrison et~al.(2016)Morrison, Law, Barrowclough, Bentall, Haddock,
  Jones, Kilbride, Pitt, Shryane, Tarrier, Welford, and
  Dunn}]{PsychologicalApproachesRecoveryBD_Morrison_16}
Anthony~P. Morrison, Heather Law, Christine Barrowclough, Richard~P. Bentall,
  Gillian Haddock, Steven~Huntley Jones, Martina Kilbride, Elizabeth Pitt,
  Nicholas Shryane, Nicholas Tarrier, Mary Welford, and Graham Dunn. 2016.
\newblock \href {https://doi.org/10.3310/pgfar04050} {{Psychological approaches
  to understanding and promoting recovery in psychosis and bipolar disorder: a
  mixed-methods approach}}.
\newblock \emph{Programme Grants for Applied Research}, 4(5):1--272.

\bibitem[{Novick et~al.(2010)Novick, Swartz, and
  Frank}]{SuicideAttemptsBP_Novick_10}
Danielle~M. Novick, Holly~A. Swartz, and Ellen Frank. 2010.
\newblock \href {https://doi.org/10.1111/j.1399-5618.2009.00786.x} {{Suicide
  attempts in bipolar I and bipolar II disorder: a review and meta-analysis of
  the evidence}}.
\newblock \emph{Bipolar disorders}, 12(1):1--9.

\bibitem[{O'Dea et~al.(2015)O'Dea, Wan, Batterham, Calear, Paris, and
  Christensen}]{SuicidalityTwitter_ODea_15}
Bridianne O'Dea, Stephen Wan, Philip~J. Batterham, Alison~L. Calear, Cecile
  Paris, and Helen Christensen. 2015.
\newblock \href {https://doi.org/10.1016/j.invent.2015.03.005} {{Detecting
  suicidality on Twitter}}.
\newblock \emph{Internet Interventions}, 2(2):183--188.

\bibitem[{O'Keeffe and
  McCarthy(2010)}]{RoutledgeHandbookCorpusLinguistics_OKeeffe_10}
Anne O'Keeffe and Michael McCarthy. 2010.
\newblock \href {https://doi.org/10.1109/IEMBS.2010.5626267} {\emph{{The
  Routledge Handbook of Corpus Linguistics}}}.
\newblock Routledge Handbooks in Applied Linguistics. Routledge.

\bibitem[{van Os et~al.(2019)van Os, Guloksuz, Vijn, Hafkenscheid, and
  Delespaul}]{EvidenceBasedPsychiatryChange_Os_19}
Jim van Os, Sinan Guloksuz, Thomas~Willem Vijn, Anton Hafkenscheid, and
  Philippe Delespaul. 2019.
\newblock \href {https://doi.org/10.1002/wps.20609} {{The evidence-based
  group-level symptom-reduction model as the organizing principle for mental
  health care: time for change?}}
\newblock \emph{World Psychiatry}, 18(1):88--96.

\bibitem[{Paul and Dredze(2017)}]{SocialMonitoringPublicHealth_Paul_17}
Michael~J. Paul and Mark Dredze. 2017.
\newblock \href {https://doi.org/10.2200/S00791ED1V01Y201707ICR060} {{Social
  Monitoring for Public Health}}.
\newblock \emph{Synthesis Lectures on Information Concepts, Retrieval, and
  Services}, 9(5):1--183.

\bibitem[{Pennebaker et~al.(2015)Pennebaker, Boyd, Jordan, and
  Blackburn}]{LIWC_Pennebaker_15}
James~W. Pennebaker, Ryan~L. Boyd, Kayla Jordan, and Kate Blackburn. 2015.
\newblock \href {https://doi.org/10.15781/T29G6Z} {{The Development and
  Psychometric Properties of LIWC2015}}.
\newblock Technical report, University of Texas at Austin, Austin.

\bibitem[{Pianta et~al.(2002)Pianta, Bentivogli, and
  Girardi}]{MultiWordNet_Pianta_02}
Emanuele Pianta, Luisa Bentivogli, and Christian Girardi. 2002.
\newblock \href {http://multiwordnet.fbk.eu/paper/MWN-India-published.pdf}
  {{MultiWordNet: developing an aligned multilingual database}}.
\newblock In \emph{Proceedings of the 1st International WordNet Conference},
  pages 293--302.

\bibitem[{Piao et~al.(2016)Piao, Rayson, Archer, Bianchi, Dayrell,
  Jim{\'{e}}nez, Knight, K{\v{r}}en, L{\"{o}}fberg, Nawab, Shafi, Teh, and
  Mudraya}]{USAS_Multilingual_Piao_16}
Scott Piao, Paul Rayson, Dawn Archer, Francesca Bianchi, Carmen Dayrell,
  Ricardo-maría Jim{\'{e}}nez, Dawn Knight, Michal K{\v{r}}en, Laura
  L{\"{o}}fberg, Muhammad~Adeel Nawab, Jawad Shafi, Phoey~Lee Teh, and Olga
  Mudraya. 2016.
\newblock {Lexical Coverage Evaluation of Large-scale Multilingual Semantic
  Lexicons for Twelve Languages}.
\newblock \emph{Tenth International Conference on Language Resources and
  Evaluation}, pages 2614--2619.

\bibitem[{Prinstein et~al.(2008)Prinstein, Nock, Simon, Aikins, Cheah, and
  Spirito}]{TrajectoriesSuicidalIdeation_Prinstein_08}
Mitchell~J. Prinstein, Matthew~K. Nock, Valerie Simon, Julie~Wargo Aikins,
  Charissa S.~L. Cheah, and Anthony Spirito. 2008.
\newblock \href {https://doi.org/10.1037/0022-006X.76.1.92.Longitudinal}
  {{Longitudinal Trajectories and Predictors of Adolescent Suicidal Ideation
  and Attempts Following Inpatient Hospitalization}}.
\newblock \emph{Journal of Consulting and Clinical Psychology}, 76(1):92--103.

\bibitem[{Rayson(2008)}]{KeyWordsKeySemanticDomainsWMatrix_Rayson_08}
Paul Rayson. 2008.
\newblock \href {https://doi.org/10.1075/ijcl.13.4.06ray} {{From key words to
  key semantic domains}}.
\newblock \emph{International Journal of Corpus Linguistics}, 13(4):519--549.

\bibitem[{Rayson(2015)}]{ComputationalToolsMethodsCorpus_Rayson_15}
Paul Rayson. 2015.
\newblock {Computational tools and methods for corpus compilation and
  analysis}.
\newblock In Douglas Biber and Randi Reppen, editors, \emph{The Cambridge
  Handbook of English corpus linguistics}, pages 32--49. Cambridge University
  Press.

\bibitem[{Rayson et~al.(2004)Rayson, Archer, Piao, and
  McEnery}]{USAS_Rayson_04}
Paul Rayson, Dawn Archer, Scott Piao, and Tony McEnery. 2004.
\newblock \href {http://eprints.lancs.ac.uk/1783/} {{The UCREL semantic
  analysis system.}}
\newblock \emph{Proceedings of the beyond named entity recognition semantic
  labelling for NLP tasks workshop}, (February 2017):7--12.

\bibitem[{Resnik et~al.(2014)Resnik, Resnik, and Mitchell}]{CLPsych1_Resnik_14}
Philip Resnik, Rebeca Resnik, and Margaret Mitchell. 2014.
\newblock \href {https://doi.org/10.1083/jcb.106.5.1795} {\emph{{Proceedings of
  the Workshop on Computational Linguistics and Clinical Psychology From
  Linguistic Signal to Clinical Reality}}}.
\newblock Association for Computational Linguistics.

\bibitem[{Russell and Browne(2005)}]{StayingWellWithBD_Russell_05}
Sarah~J. Russell and Jan~L. Browne. 2005.
\newblock {Staying well with bipolar}.
\newblock \emph{Australian {\&} New Zealand Journal of Psychiatry},
  39(3):187--193.

\bibitem[{Sale et~al.(2002)Sale, Lohfeld, and
  Brazil}]{MixedMethodsHealthCare_Sale_02}
Joanna E.~M. Sale, Lynne~H. Lohfeld, and Kevin Brazil. 2002.
\newblock \href {https://doi.org/10.1023/A:1014301607592} {{Revisiting the
  Quantitative-Qualitative Debate: Implications for Mixed-Methods Research}}.
\newblock \emph{Quality {\&} Quantity}, 36:43--53.

\bibitem[{Sanches and Jorge(2004)}]{TransculturalBD_Sanches_04}
Marsal Sanches and Miguel~Roberto Jorge. 2004.
\newblock {Transcultural aspects of bipolar disorder}.
\newblock \emph{Brazilian Journal of Psychiatry}, 26(3):54--56.

\bibitem[{Schwartz et~al.(2017)Schwartz, Giorgi, Sap, Crutchley, Eichstaedt,
  and Ungar}]{DLATK_Schwartz_17}
H.~Andrew Schwartz, Salvatore Giorgi, Maarten Sap, Patrick Crutchley,
  Johannes~C. Eichstaedt, and Lyle Ungar. 2017.
\newblock \href {https://doi.org/10.18653/v1/d17-2010} {{DLATK: Differential
  Language Analysis ToolKit}}.
\newblock In \emph{Proceedings of the 2017 EMNLP System Demonstrations}, pages
  55--60.

\bibitem[{Sekuli{\'{c}} et~al.(2018)Sekuli{\'{c}}, Gjurkovi{\'{c}}, and
  {\v{S}}najder}]{BDPrediction_Sekulic_18}
Ivan Sekuli{\'{c}}, Matej Gjurkovi{\'{c}}, and Jan {\v{S}}najder. 2018.
\newblock \href {https://doi.org/10.18653/v1/P17} {{Not Just Depressed: Bipolar
  Disorder Prediction on Reddit}}.
\newblock In \emph{WASSA@EMNLP}, 2001, pages 72--78, Brussels. Association for
  Computational Linguistics.

\bibitem[{Semino et~al.(2017)Semino, Demj{\'{e}}n, Hardie, Payne, and
  Rayson}]{MetaphorCancer_Semino_17}
Elena Semino, Zsófia Demj{\'{e}}n, Andrew Hardie, Sheila Payne, and Paul
  Rayson. 2017.
\newblock \href {https://doi.org/10.4324/9781315629834} {\emph{{Metaphor,
  Cancer and the End of Life: A Corpus-Based Study}}}.

\bibitem[{Slade et~al.(2012)Slade, Leamy, Bacon, Janosik, Le~Boutillier,
  Williams, and
  Bird}]{InternationalDifferencesRecoverySystematicReview_Slade_12}
M.~Slade, M.~Leamy, F.~Bacon, M.~Janosik, C.~Le~Boutillier, J.~Williams, and
  V.~Bird. 2012.
\newblock \href {https://doi.org/10.1017/S2045796012000133} {{International
  differences in understanding recovery: Systematic review}}.
\newblock \emph{Epidemiology and Psychiatric Sciences}, 21(4):353--364.

\bibitem[{Steckler et~al.(1992)Steckler, McLeroy, Goodman, Bird, and
  McCormick}]{IntegratingQualiQuanti_Steckler_92}
Allan Steckler, Kenneth~R. McLeroy, Robert~M. Goodman, Sheryl~T. Bird, and
  Lauri McCormick. 1992.
\newblock \href {https://doi.org/10.1177/109019819201900101} {{Toward
  Integrating Qualitative and Quantitative Methods: An Introduction}}.
\newblock \emph{Health Education {\&} Behavior}, 19(1):1--8.

\bibitem[{Strakowski et~al.(1998)Strakowski, Keck, McElroy, West, Sax, Hawkins,
  Kmetz, Upadhyaya, Tugrul, and
  Bourne}]{OutcomeAffectivePsychosis_Strakowski_98}
Stephen~M. Strakowski, Paul~E. Keck, Susan~L. McElroy, Scott~A. West, Kenji~W.
  Sax, John~M. Hawkins, Geri~F. Kmetz, Vidya~H. Upadhyaya, Karen~C. Tugrul, and
  Michelle~L. Bourne. 1998.
\newblock \href {https://doi.org/10.1001/archpsyc.55.1.49} {{Twelve-Month
  Outcome After a First Hospitalization for Affective Psychosis}}.
\newblock \emph{Archives of General Psychiatry}, 55(1):49--55.

\bibitem[{Stuart et~al.(2017)Stuart, Tansey, and
  Quayle}]{RecoverySystematicReview_Stuart_17}
Simon~Robertson Stuart, Louise Tansey, and Ethel Quayle. 2017.
\newblock \href {https://doi.org/10.1080/09638237.2016.1222056} {{What we talk
  about when we talk about recovery: a systematic review and best-fit framework
  synthesis of qualitative literature}}.
\newblock \emph{Journal of Mental Health}, 26(3):291--304.

\bibitem[{Tackman et~al.(2018)Tackman, Sbarra, Carey, Donnellan, Horn,
  Holtzman, Edwards, Pennebaker, and Mehl}]{DepressionMultiLab_Tackman_18}
Allison~M. Tackman, David~A. Sbarra, Angela~L. Carey, M.~Brent Donnellan,
  Andrea~B. Horn, Nicholas~S. Holtzman, To'Meisha~S. Edwards, James~W.
  Pennebaker, and Matthias~R. Mehl. 2018.
\newblock \href {https://doi.org/10.1037/pspp0000187} {{Depression, Negative
  Emotionality, and Self-Referential Language: A Multi-Lab, Multi-Measure, and
  Multi-Language-Task Research Synthesis}}.
\newblock \emph{Journal of Personality and Social Psychology}, (March).

\bibitem[{Tashakkori and Teddlie(1998)}]{MixedMethods_Tashakkori_98}
Abbas Tashakkori and Charles Teddlie. 1998.
\newblock \emph{{Mixed methodology: Combining qualitative and quantitative
  approaches}}, volume~46.
\newblock Sage.

\bibitem[{Tohen et~al.(2003)Tohen, Zarate, Hennen, Khalsa, Strakowski,
  Gebre-Medhin, Salvatore, and Baldessarini}]{FirstEpisodeManiaStudy_Tohen_03}
Mauricio Tohen, Carlos~A. Zarate, John Hennen, Hari Mandir~Kaur Khalsa,
  Stephen~M. Strakowski, Priscilla Gebre-Medhin, Paola Salvatore, and Ross~J.
  Baldessarini. 2003.
\newblock \href {https://doi.org/10.1176/appi.ajp.160.12.2099} {{The
  McLean-Harvard first-episode mania study: Prediction of recovery and first
  recurrence}}.
\newblock \emph{American Journal of Psychiatry}, 160(12):2099--2107.

\bibitem[{Trampus(2015)}]{TwitterLanguageIdentification_15}
Mitja Trampus. 2015.
\newblock \href
  {https://blog.twitter.com/2015/evaluating-language-identification-performance}
  {{Evaluating language identification performance}}.

\bibitem[{Turney and Pantel(2010)}]{FrequencyToMeaning_Turney_10}
Peter~D. Turney and Patrick Pantel. 2010.
\newblock {From Frequency to Meaning: Vector Space Models of Semantics}.
\newblock \emph{Journal of Artificial Intelligence Research}, 37:141--188.

\bibitem[{{U.S. Department of Health and Human Services: The National Institute
  of Mental Health}(2016)}]{Schizophrenia_NIMH_16}
{U.S. Department of Health and Human Services: The National Institute of Mental
  Health}. 2016.
\newblock \href
  {https://www.nimh.nih.gov/health/topics/schizophrenia/index.shtml}
  {{Schizophrenia}}.

\bibitem[{Wang and Jurgens(2018)}]{MeasuringSupportOnlineCommunities_Wang_18}
Zijian Wang and David Jurgens. 2018.
\newblock \href {http://www.aclweb.org/anthology/D18-1004} {{It's going to be
  okay: Measuring Access to Support in Online Communities}}.
\newblock In \emph{Proceedings of the 2018 Conference on Empirical Methods in
  Natural Language Processing (EMNLP)}, pages 33--45.

\bibitem[{Williams et~al.(2017)Williams, Burnap, and
  Sloan}]{EthicalFrameworkPublishingTwitterData_Williams_17}
Matthew~L. Williams, Pete Burnap, and Luke Sloan. 2017.
\newblock \href {https://doi.org/10.1177/0038038517708140} {{Towards an Ethical
  Framework for Publishing Twitter Data in Social Research: Taking into Account
  Users’ Views, Online Context and Algorithmic Estimation}}.
\newblock \emph{Sociology}, 51(6):1149--1168.

\bibitem[{Young and Garett(2018)}]{SuicdeOnlineStudy_Young_18}
Sean~D. Young and Renee Garett. 2018.
\newblock \href {https://doi.org/10.2196/mental.8971} {{Ethical issues in
  addressing social media posts about suicidal intentions during an online
  study among youth: case study}}.
\newblock \emph{Journal of Medical Internet Research}, 20(5):1--5.

\bibitem[{Young and Ensing(1999)}]{ExploringRecovery_Young_99}
Sharon~L. Young and David~S. Ensing. 1999.
\newblock \href
  {http://ezproxy.deakin.edu.au/login?url=http://search.ebscohost.com/login.aspx?direct=true&db=cinref&AN=PRJ.BB.BAI.YOUNG.ERFPPP&site=ehost-live}
  {{Exploring recovery from the perspective of people with psychiatric
  disabilities.}}
\newblock \emph{Psychiatric Rehabilitation Journal}, 22(3):219--231.

\bibitem[{Zeman et~al.(2018)Zeman, Hajic, Popel, Straka, Nivre, Ginter, Petrov,
  and Oepen}]{CoNLLST_Zeman_18}
Daniel Zeman, Jan Hajic, Martin Popel, Milan Straka, Joakim Nivre, Filip
  Ginter, Slav Petrov, and Stephan Oepen. 2018.
\newblock {Proceedings of the CoNLL 2018 Shared Task: Multilingual Parsing from
  Raw Text to Universal Dependencies}.
\newblock In \emph{The SIGNLL Conference on Computational Natural Language
  Learning}.

\end{thebibliography}
\bibliographystyle{acl_natbib}

\appendix



\end{document}